\documentclass[letterpaper, 10 pt, conference]{ieeeconf}  

\IEEEoverridecommandlockouts                              

\usepackage[T1]{fontenc}
\usepackage{graphicx}
\usepackage{subfigure}
\usepackage{booktabs}
\usepackage{amsmath}
\usepackage{amsfonts}
\usepackage{algorithm}
\usepackage{algorithmic}
\usepackage{balance}
\usepackage{hyperref}

\title{\LARGE \bf
Policy Constraint by Only Support Constraint for Offline Reinforcement Learning
}

\author{Yunkai Gao$^{1}$, Jiaming Guo$^{3}$, Fan Wu$^{2}$ and Rui Zhang$^{3}$*
\thanks{$^{1}$Y. Gao is with the University of Science and Technology of China.
        {\tt\small gyk314@mail.ustc.edu.cn}}%
\thanks{$^{2}$W. Fan is with the Intelligent Software Research Center, Institute of Software, CAS.
        {\tt\small wufan2020@iscas.ac.cn}}%
\thanks{$^{3}$J. Guo, R. Zhang are with SKL of Processors, Institute of Computing Technology, CAS.
        {\tt\small zhangrui@ict.ac.cn, guojiaming@ict.ac.cn}}%
\thanks{*Corresponding author: Rui Zhang}%
}

\begin{document}

\maketitle
\thispagestyle{empty}
\pagestyle{empty}

\begin{abstract}
Offline reinforcement learning (RL) aims to optimize a policy by using pre-collected datasets, to maximize cumulative rewards.
However, offline reinforcement learning suffers challenges due to the distributional shift between the learned and behavior policies, leading to errors when computing Q-values for out-of-distribution (OOD) actions.  
To mitigate this issue, policy constraint methods aim to constrain the learned policy's distribution with the distribution of the behavior policy or confine action selection within the support of the behavior policy. 
However, current policy constraint methods tend to exhibit excessive conservatism, hindering the policy from further surpassing the behavior policy's performance. 
In this work, we present Only Support Constraint (OSC) which is derived from maximizing the total probability of learned policy in the support of behavior policy, to address the conservatism of policy constraint. 
OSC presents a regularization term that only restricts policies to the support without imposing extra constraints on actions within the support.
Additionally, to fully harness the performance of the new policy constraints, OSC utilizes a diffusion model to effectively characterize the support of behavior policies.
Experimental evaluations across a variety of offline RL benchmarks demonstrate that OSC significantly enhances performance, alleviating the challenges associated with distributional shifts and mitigating conservatism of policy constraints.
Code is available at \href{https://github.com/MoreanP/OSC}{https://github.com/MoreanP/OSC}.
\end{abstract}

\section{INTRODUCTION}

Reinforcement learning (RL) has achieved great success in many decision-making tasks \cite{dqn,master_alpha_go}.
However, online RL needs to interact with the environment during training, which limits its application in some fields, such as autonomous driving\cite{driving}, medical healthcare \cite{healthcare}, and robot control\cite{robot}, because of the high cost and danger of interacting with these environments.
To solve this problem, offline RL\cite{offline_rl_survey} learns on pre-collected offline datasets without additional interaction with the environment during training.
The off-policy RL methods can be applied in offline RL. 
However, the evaluation of policies requires querying the $Q$-function of the actions derived from the learned policy.
Due to the distribution shift between the learned policy and the behavior policy, certain actions might not be in the offline datasets.
The resultant extrapolation error in the $Q$-function, stemming from these out-of-distribution (OOD) actions, can potentially overestimate subsequent $Q$-function values.
This precipitates training instabilities \cite{offline_rl_survey, bcq}.

Many offline RL methods have recently been proposed to address the distribution shift problem.
The main approach is to introduce conservatism into offline RL algorithms to ensure that the learned policy remains within the offline dataset distribution. 
Policy constraint methods use divergence constraints \cite{brac,fisher-brc} like KL divergence to confine distributions of the learned policy and behavior policy closer together.
Aside from that, another type of policy constraint method \cite{bear, spot, cped} seeks to directly restrict the learned policy to the support of the behavior policy.
These methods encourage the learned policy to select actions similar to the offline datasets, thereby reducing the negative impact of OOD actions.
However, the current policy constraint methods tend to be overly conservative. 
The specific probabilities of the behavior policy influence the strength of these constraint terms.
These constraint terms apply varying degrees of constraint across different actions, with stronger constraints on high-probability actions and weaker ones on low-probability actions. 
When the behavior policy's performance is poor and high-quality actions have low probabilities in the support, these constraint terms can trap the learned policy in high-probability but low-quality actions.
This hinders the learned policy from further improving performance.


In this work, We introduce \textbf{O}nly \textbf{S}upport \textbf{C}onstraint (\textbf{OSC}) to alleviate the conservatism of constraints.
The core idea of OSC is that no additional varying constraints should be imposed on actions.
Concretely, OSC only constrains the learned policy's actions within the range of the support through the regularization term.
Notably, OSC refrains from imposing additional constraints on actions that already fall within the support, so that the learned policy can freely choose the better action within the confines of the support. 
OSC starts from the maximization of the learned policy’s total probability within the support of the behavior policy, obtaining a new constraint regularization term.
However, the new constraint term requires a more accurate estimation of the support. 
To fully leverage the performance of the new constraint term, OSC utilizes the diffusion model \cite{diffusion, diffusionql} to explicitly model the extent of the behavioral policy's support.
We widely validate the effectiveness of OSC on the D4RL benchmark datasets which are widely used by prior offline RL methods.

To summarize, the contributions of this paper are as follows:
\begin{itemize}
    \item We obtain a new regularization term of support constraint from the total probability that the learned policy resides in the support of behavior policy.
    \item We propose \textbf{O}nly \textbf{S}upport  \textbf{C}onstraint (OSC) to implement the regularization term by using the diffusion model to model the support of behavior policy.
    \item Compared with the previous offline RL methods, OSC achieves SOTA results on the benchmark datasets.
\end{itemize}

\section{Related Work}

The extrapolation error resulting from the distribution shift between the learned policy and the behavior policy often leads to the failure of most online off-policy methods in offline reinforcement learning (RL).
As a result, the majority of offline RL approaches build upon the foundation of off-policy methods and introduce constraint terms to encourage proximity between the learned policy and the behavior policy \cite{brac, fisher-brc, bear, spot, csro}.
Other methods have also been employed to address this issue: uncertainty estimation \cite{rorl,sac-n,pbrl}, 
conservative value estimation \cite{cql, mcq, pbrl}, 
and in-sample methods \cite{iql, sql, inac}.
Our method is a form of policy constraint method, and we review previous instances of policy constraint methods.

\paragraph{Policy constraint methods.} 
Prior policy constraint methods aimed to confine the learned policy closer to the behavior policy, mitigating the estimation error of out-of-distribution (OOD) actions caused by the distributional dissimilarity:
BCQ \cite{bcq} modeled the learned policy as a perturbation on top of the behavior policy, employing a Conditional Variational Autoencoder (CVAE) \cite{cvae} to represent the behavior policy and utilizing a maximum value constraint for perturbation training. 
To confine the learned policy within the support of the behavior policy, BEAR \cite{bear} employed Maximum Mean Discrepancy (MMD) as an approximation for support constraint.
BRAC \cite{brac}, on the other hand, directly imposed constraint terms during policy estimation and updates, such as KL divergence, MMD constraint, and Wasserstein constraint.
TD3+BC\cite{td3+bc} took a simpler approach by building upon TD3 and adding a maximum likelihood estimate of behavior cloning (BC) loss as a regularization term. 
SPOT \cite{spot}, departing directly from probability density in the support, introduced novel regularization terms and employed a CVAE to model the density of the behavior policy.
Due to the excessively conservative current policy constraint methods, we proposed OSC which is a novel support constraint method.

\paragraph{Diffusion models in RL.}
The denoising diffusion probabilistic model (Diffusion) \cite{diffusion} formulates the generation process as an MDP process tied to noise, divided into a forward process gradually introducing noise to the original distribution and a reverse process reconstructing the original distribution from noise. 
This empowers the diffusion model with a stronger ability to fit arbitrary distributions. 
In Diffusion-QL \cite{diffusionql}, the learned policy is modeled in the form of diffusion, employing a TD3-BC style algorithm that uses the loss from behavior cloning of the diffusion model as the BC constraint term. 
Diffuser \cite{diffuser} employs the diffusion model to directly construct the distribution of trajectories rather than the distribution of transition pairs. 
It further trains a return model to predict the cumulative reward of trajectories generated by the diffusion trajectory generator. 
SfBC \cite{sfbc}, in contrast, involves constructing the behavior policy using the diffusion approach and then performing resampling using Q-value to weighted actions sampled from the behavior policy.
AdaptDiffuser \cite{adaptdiffuser} introduces a diffusion model to generate expert trajectories, then selects high-quality trajectories via a reward-guided discriminator to improve the generalization ability.

\section{Preliminaries}
\subsection{Offline Reinforcement Learning}
We consider the RL problem as a Markov Decision Process (MDP), defined as a tuple $M=\langle \mathcal{S},\mathcal{A}, T(s'|s, a),r(s, a),\rho(s_{0}),\gamma\rangle$, consisting of state space $\mathcal{S}$, action space $\mathcal{A}$, transition distribution function $T(s'|s, a)$, reward function $r(s, a)$, initial state distribution $\rho(s_{0})$, and discount factor $\gamma \in (0, 1)$.

The goal of RL is to train a learned policy $\pi_{\theta}(a|s)$ that maximizes the expected cumulative rewards $J(\pi_{\theta})=\mathbb{E}_{\tau}[\sum_{t=0}^{\infty}\gamma^{t}R(s_{t},a_{t})]$, where $\tau$ is trajectory following $s_{0}\sim \rho_{0}$, $a_{t}\sim \pi(a_{t}|s_{t})$, $s_{t+1}\sim T(s_{t+1}|s_{t}, a_{t})$.

Under the frameworks of Actor-Critic, the optimization objectives for policy evaluation and policy updates are, respectively:
\begin{equation}\label{eq:qloss}
    \begin{aligned}
        L_{Q}(\psi)=&\mathbb{E}_{(s,a,r,s')\sim D,a'\sim\pi_{\theta}(\cdot|s')} \\
        &[Q_{\psi}(s,a) -r-\gamma Q_{\bar{\psi}}(s',a')]^{2}
    \end{aligned}
\end{equation}
\begin{equation}\label{eq:dpg_ploss}
    L_{\pi}(\theta)=\mathbb{E}_{s\sim D,a\sim \pi_{\theta}(\cdot|s)}[-Q_{\psi}(s,a)]
\end{equation}

Unlike online RL methods that can interact with the environment to collect experience data, offline RL employs a fixed dataset $D=\{(s, a,r,s')\}$ pre-collected using an unknown behavior policy $\pi_{\beta}(a|s)$ for training. 
Applying off-policy methods directly to offline RL becomes challenged by the $Q$ estimation errors introduced by out-of-distribution actions.
This is because when optimizing the learned policy by maximizing the $Q$ function, the $Q$ function may overestimate some OOD actions, and the learned policy tends to choose these actions.
However, in offline settings, the learned policy cannot correct the overestimation of $Q$ by interacting with the environment to obtain new data. 
The error of $Q$ will be transmitted throughout the entire training process, leading to training failure.

\subsection{Diffusion Model}


Diffusion-based generative model\cite{diffusion} contains a forward noising process and a backward denoising process.
In the forward process, Gaussian noise is added to origin data $x_{0}$ over $T$ steps, generating a sequence of $x_{1:T}$ until it nearly becomes pure Gaussian noise.
The forward process follows a variance schedule $\beta_{1:T}$, $0<\beta_{1}<\beta_{2}<\cdots<\beta_{T}<1$. 
The relation between $x_{t-1}$ and $x_{t}$ is $x_{t}=\sqrt{1-\beta_{t}}x_{t-1}+\beta_{t}z_{t}$, where $z_{t}\sim\mathcal{N}(0,I)$.
After noising $T-1$ times, the relation between $x_{0}$ and $x_{t}$ is $x_{t}=\sqrt{\bar{\alpha}_{t}}x_{0}+\sqrt{1-\bar{\alpha}_{t}}z_{t}$, where $\alpha_{t}=1-\beta_{t}$ and $\bar{\alpha}_{t}=\alpha_{1}\alpha_{2}\dots\alpha_{t}$.
The reverse denoising process is constructed as $p(x_{0:T})\sim\mathcal{N}(x_{t};0,I)\prod_{t=1}^{T}p(x_{t-1}|x_{t})$. Through the Bayesian equation, 
\begin{equation*}\label{eq:diffusion_back}
    \begin{aligned}
        p(x_{t-1}|x_{t})&\sim\mathcal{N}(\widetilde{\mu}_{t};\widetilde{\beta}_{t}) \\
        \widetilde{\mu}_{t}=\frac{1}{\sqrt{\alpha_{t}}}(x_{t}-\frac{1-\alpha_{t}}{\sqrt{1-\bar{\alpha}_{t}}}&\epsilon_{\phi}), \widetilde{\beta}_{t}=\frac{1-\bar{\alpha}_{t-1}}{1-\bar{\alpha}_{t}}\beta_{t}.
    \end{aligned}
\end{equation*}
The optimization object is maximizing the evidence lower bound, the corresponding loss is 
\begin{equation*}\label{eq:diffusion_loss}
    L_{\phi}=\mathbb{E}_{x_{0},\epsilon}[\Vert\epsilon-\epsilon_{\phi}(\sqrt{\bar{\alpha}_{t}}x_{0}+\sqrt{1-\bar{\alpha}_{t}}\epsilon, t)]\Vert^{2}.
\end{equation*}

\section{Method}
\begin{figure}[t]
    \centering
    \includegraphics[width=0.9\linewidth]{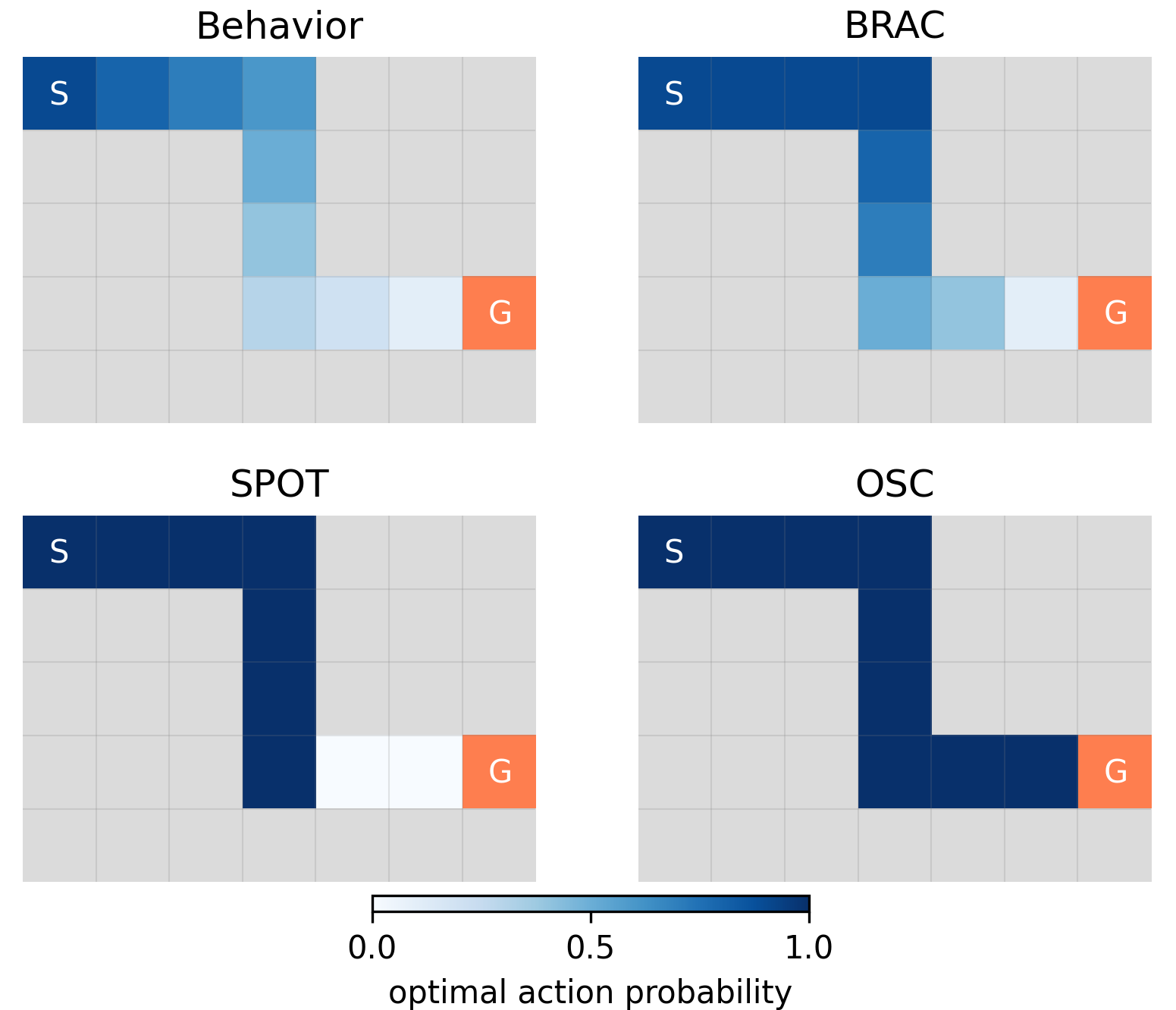}
    \caption{
    Visualizing the impact of excessive conservatism.
    The environment is a grid task, "S" is the start location, and "G" is the goal location of the agent.
    }
    \label{fig:demo}
\end{figure}

In this section, we first use a motivating example to illustrate why the existing policy constraints as previously mentioned suffer from excessive conservatism.
Therefore we propose a simple yet effective method, called \textbf{O}nly \textbf{S}upport \textbf{C}onstraint (\textbf{OSC}), a policy constraint method that addresses current conservatism through support constraints.
Specifically, we use a support constraint based on the probability of learned policy within the support of behavior policy.
This only limits the learned policy to the support of behavior policy but does not impose extra constraints on actions within the support.
We employ the diffusion model to enhance the accuracy of estimating the behavior policy's support. 
This model effectively constructs the behavior policy's support.

\subsection{Motivation Example}

We use a simple gird task to demonstrate the excessive conservatism problem as shown in Fig. \ref{fig:demo}.
The agent needs to navigate to the bottom-right "G" from the top-left "S".
The episode is considered terminated when the agent reaches "G" or enters a gray area.
The agent in this task has only 2 actions, including "right" and "down". 
This means that the agent can only move along a "z" shaped trajectory, as shown in the blue trajectory in the figure.
The agent only receives a reward of 1 upon reaching "G" and receives a reward of 0 in all other locations.
The 'Behavior' illustrates the action probabilities of the behavior policy, with the probability of selecting the optimal action decreasing as the agent gets closer to "G".
The probability of selecting the optimal action is visualized from 0 to 1, represented by shades from white to deep blue.
For offline training, we use this behavior policy to collect offline datasets consisting of one million steps.

We evaluate BRAC\cite{brac} and SPOT\cite{spot} on this task. 
We performed a discretization process similar to discrete-SAC ~\cite{discrete-sac} to adapt these methods to discrete environments. The policy $\pi(s)$, action-value function $Q(s)$, and behavior policy $\pi_{\beta}(s)$ output $Q$-values or action probabilities for all discrete actions. 
We visualized the probability of selecting the optimal action for the learned policy in Fig. \ref{fig:demo}.
We observed that as the probability of selecting the optimal action under the behavior policy decreases, the learning policy of BRAC exhibits a similar trend. 
When the probability of the optimal action in the behavior policy is relatively high, SPOT \cite{spot} performs well in choosing the better actions. 
However, SPOT tends to break down as it becomes heavily constrained to higher-probability poor action.
The constraint term for BRAC is $\mathbb{E}_{a\sim\pi_{\theta}(\cdot|s)}[\log\pi_{\theta}(a|s)-\log\pi_{\beta}(a|s)]$ and the constraint term for SPOT is $\mathbb{E}_{a\sim\pi_{\theta}(\cdot|s)}[-\log\pi_{\beta}(a|s)]$.
we can observe that the constraint magnitude varies for different actions. When the behavior policy has a high probability of selecting suboptimal actions, the constraint becomes even stronger.
This ultimately results in the policy constraint being overly conservative.
Hence, this example motivates us to propose a new policy constraint term.

\subsection{Support Constraint via Behavior Density}
When the probability of an action according to the behavior policy is 0 or too small, the action is rarely observed within the offline datasets.
The $Q$-function will encounter substantial errors while estimating the value of such actions.
~\cite{bear, spot} define the support of the behavior policy on conditioned $s$ as $\{a\in A| \pi_{\beta}(a|s)>\epsilon\}$ and introduce the support operator:
\begin{equation}\label{eq:support_bellman}
    \mathcal{T}_{\epsilon}Q(s,a)=\mathbb{E}_{s'}[r+\gamma \max\limits_{a':\pi_{\beta}(a'|s')>\epsilon}Q(s',a')]
\end{equation}
The fixed point $Q_{\epsilon}^{*}$ is named as the supported optimal $Q$-function.
Different from common policy extraction, the policy extraction of support optimization needs to extract the optimal policy within the support: 
\begin{equation}\label{eq:support_maxQ}
    \pi_{\epsilon}^{*}(s) = \arg \max\limits_{a:\pi_{\beta}(a|s)>\epsilon} Q_{\epsilon}^{*}(s,a).
\end{equation}

The previous analysis mentioned that the constraint terms in BRAC and SPOT impose varying degrees of constraints on different actions, which is the reason for the conservative.
According to Eq. \ref{eq:support_maxQ}, an ideal support constraint only confines the learned policy within the boundaries of the support. 
Within this support, there should be no imposition of extra constraints. 
This approach ensures that the policy can opt for better action in the support without extra limitations.
However, none of BRAC and SPOT have fully conformed to the ideal form presented in Eq. \ref{eq:support_maxQ}.

To solve this problem, we proposed OSC. 
We start from the probability of behavior policy $\pi_{\beta}(\cdot|s)$, and maximize the probability of learned policy $\pi_{\theta}(\cdot|s)$ within the support of behavior policy $\pi_{\beta}(\cdot|s)$:
\begin{equation}\label{eq:support1}
    \max\limits_{\theta}\mathbb{E}_{s\sim D}[\int_{a\in A,\pi_{\beta}(a|s)>\epsilon}\pi_{\theta}(a|s)da].
\end{equation}  
We extract the part of integrating learned policy density and get:
\begin{equation}\label{eq:support2}
    \max\limits_{\theta}\mathbb{E}_{s\sim D,a\sim \pi_{\theta}(\cdot|s)}[\mathbb{I}(\log\pi_{\beta}(a|s)>\hat{\epsilon})],
\end{equation}
where $\hat{\epsilon}=\log\epsilon$, $\mathbb{I}$ is the indicator function:
\begin{equation*}
    \mathbb{I}(x)=\begin{cases}
        1 & x\; is\; true \\
        0 & x\; is\; false.
    \end{cases}
\end{equation*}
Constraints starting from the probability of learned policy in the support are intuitive. 
For mathematical convenience, we use the log-likelihood to replace the probability density of the behavior policy.

By converting the constraint of Eq. \ref{eq:support2} into a regularization term direction, combined with Eq. \ref{eq:support_maxQ}, we finally get the policy learning objective of OSC:
\begin{equation}\label{eq:ploss1}
    \begin{aligned}
        L_{\pi}(\theta)=\mathbb{E}_{s\sim D,a\sim \pi_{\theta}(\cdot|s)}[-Q_{\psi}(s,a)-\lambda\mathbb{I}(\log\pi_{\beta}(a|s)>\hat{\epsilon})],
    \end{aligned}
\end{equation}
where $\lambda$ is a hyperparameter. 
As shown in Eq. \ref{eq:ploss1}, when the learned policy $\pi_{\theta}$ is outside the support, there will be a constraint item of size $\lambda$, which limits the $\pi_{\theta}$ to the support; when it is inside the support, there is no constraint, and $\pi_{\theta}$ can be free select the better action in the support.
In this way, I can eliminate the conservatism in the divergence-based policy constraints and the current support constraints.

\begin{figure}[t]
    \centering
    \includegraphics[width=0.85\linewidth]{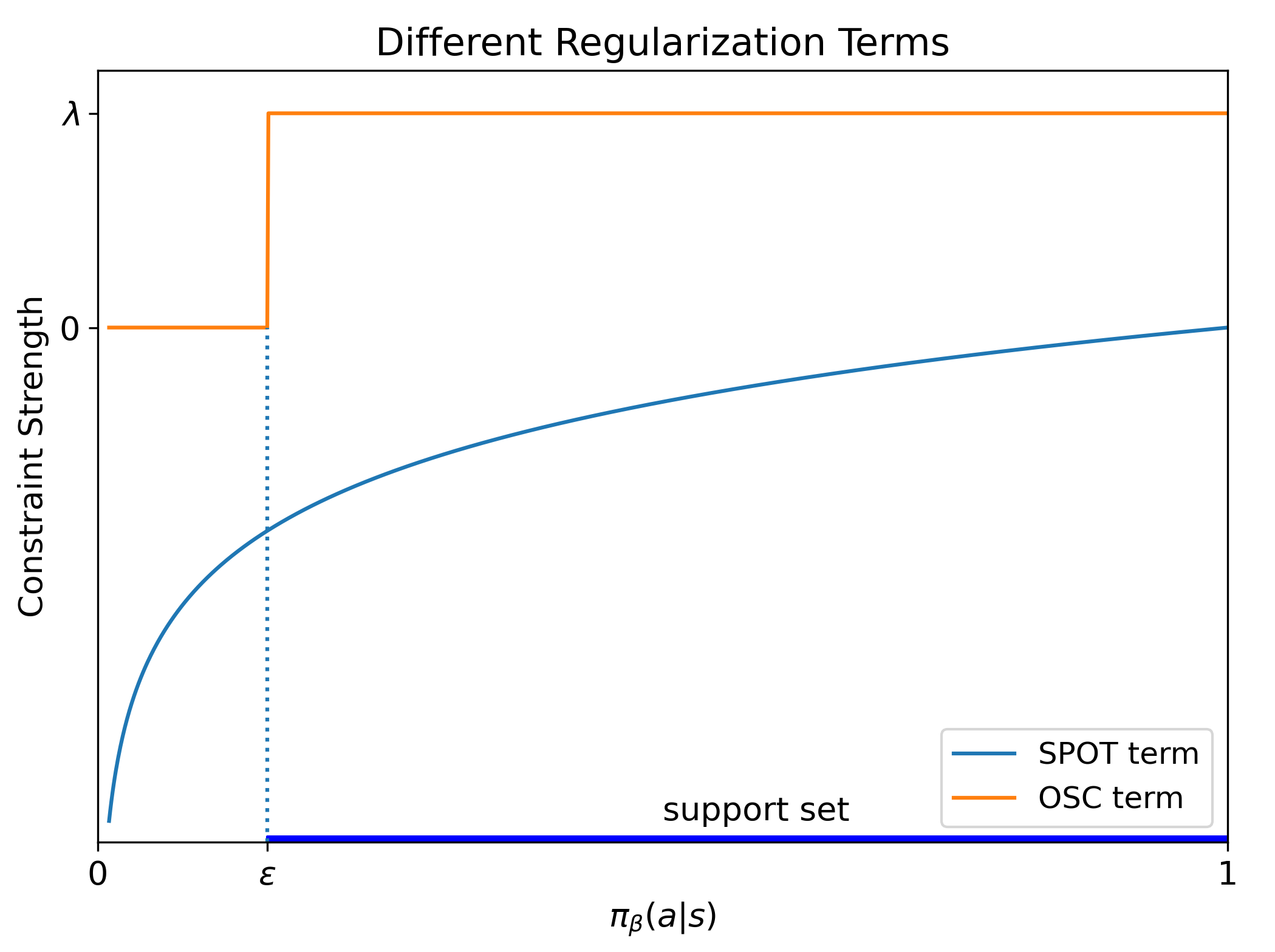}
    \caption{
    We visualize the curve graph, which shows the variation of SPOT \cite{spot} and OSC(ours) constraint terms with $\pi_{\beta}(a|s)$. 
    The blue area on the coordinate axis indicates the defined support of behavior policy, $\epsilon$ is the lower bound of the support, and $\lambda$ is the constraint strength.
    The optimization objective of SPOT and OSC is to maximize the constraint item, that is, the constraint is in the support.
    }
    \label{fig:penalty}
\end{figure}


\subsection{Estimation of Support Set}
The optimization objective shown by Eq. \ref{eq:ploss1} requires us to estimate the behavior policy $\pi_{\beta}$ from the offline datasets.
As shown in Fig. \ref{fig:penalty}, by comparing the constraint term of our optimization objective with the constraint term of SPOT, we can observe that: 
In our optimization objective, the constraint term is a mutation near the support boundary $\epsilon$. 
This leads to estimation errors near the support boundary $\epsilon$ can cause our constraint terms to suddenly change from 0 to $\lambda$ or from $\lambda$ to 0.
On the other hand, the constraint term of SPOT is linearly changing near the support boundary $\epsilon$, so the estimation error will not cause significant changes in the constraint terms.
This means that our OSC needs a more accurate estimate of the density of the $\pi_{\beta}$ to be able to accurately impose the regularization term compared to SPOT.

For a behavior policy $\pi_{\beta}$, employing the conditional variational autoencoder (CVAE) \cite{cvae} to estimate the probability density results in considerable errors.
Because diffusion fits arbitrary distributions better, it can estimate $\pi_{\beta}$ more accurately.
We train a conditional diffusion model to estimate $\pi_{\beta}$ by optimizing the variational upper bound of the negative log-likelihood $-\log \pi_{\beta}$ which is optimized by minimizing:
\begin{equation}\label{eq:diffusion_loss}
    \begin{aligned}
    L_{\pi_{\beta}}(\phi)=&\mathbb{E}_{t\sim\mathcal{U}(1,T),\epsilon\sim\mathcal{N}(0,I),(s,a)\sim \mathcal{D}}\\
    &[\Vert\epsilon-\epsilon_{\phi}(\sqrt{\hat{\alpha}_{t}}a+\sqrt{1-\hat{\alpha}_{t}}\epsilon,s,t)\Vert^{2}] \\
    \approx& -\log \pi_{\beta}(a|s)  \\
    \stackrel{\text{def}}{=}&\mathcal{F}(a|s;\phi),
    \end{aligned}
\end{equation}
where $\mathcal{U}$ is a uniform discrete distribution.

\begin{table*}[ht]
\centering
\caption{Normalized score of OSC and prior methods on MuJoCo and AntMaze datasets. m-e = "medium-expert", m = "medium", m-r = "medium-replay". 
For OSC, we report the mean and standard deviation for 10 seeds.}
\resizebox{1\linewidth}{!}{
\begin{tabular}{l||c|c|c|c|c|c|c|c}
\toprule[1pt]
Dataset             & BC   & BCQ  & DT      & TD3+BC& CQL   & IQL   & SPOT  & OSC(Ours)                 \\
\hline
halfcheetah-m-e-v2  & 55.2 & 89.1 & 86.8    & 90.7  & \textbf{91.6}  & 86.7  & 86.9  & 89.4$\pm$3.7              \\
hopper-m-e-v2       & 52.5 & 81.8 & \textbf{107.6}   & 98.0  & 105.4 & 91.5  & 99.3  & 107.0$\pm$5.1             \\
walker-m-e-v2       & 107.5& 109.0& 108.1   & 110.1 & 108.8 & 109.6 & 112.0 & \textbf{117.7}$\pm$1.4    \\
halfcheetah-m-v2    & 42.6 & 47.0 & 42.6    & 48.3  & 44.0  & 47.4  & 58.4  & \textbf{65.6}$\pm$1.0     \\
hopper-m-v2         & 52.9 & 56.7 & 67.6    & 59.3  & 58.5  & 66.2  & 86.0  & \textbf{100.9}$\pm$1.8    \\
walker-m-v2         & 75.3 & 72.6 & 74.0    & 83.7  & 72.5  & 78.3  & 86.4  & \textbf{88.9}$\pm$0.8     \\
halfcheetah-m-r-v2  & 36.6 & 40.4 & 36.6    & 44.6  & 45.5  & 44.2  & 52.2  & \textbf{55.9}$\pm$2.1     \\
hopper-m-r-v2       & 18.1 & 53.3 & 82.7    & 60.9  & 95.0  & 94.7  & 100.2 & 99.8$\pm$1.2              \\
walker-m-r-v2       & 26.0 & 52.1 & 66.6    & 81.8  & 77.2  & 73.8  & 91.6  & \textbf{93.0}$\pm$3.8     \\
\hline
Gym-MuJoCo sum     & 466.7& 602.0& 672.6    & 677.4 & 698.5 & 692.4 & 773.0 & \textbf{818.2}$\pm$20.9   \\
\hline
antmaze-umaze-v2            & 49.2 & 78.9 & 54.2 & 73.0  & 82.6 & 89.6         & 93.5          & \textbf{94.4}$\pm$3.8     \\
antmaze-umaze-diverse-v2    & 41.8 & 55.0 & 41.2 & 47.0  & 10.2 & \textbf{65.6}& 40.7          & 55.2$\pm$14.3             \\
antmaze-medium-play-v2      & 0.4  & 0.0  & 0.0  & 0.0   & 59.0 & 76.4         & 74.7          & \textbf{77.5}$\pm$5.5     \\
antmaze-medium-diverse-v2   & 0.2  & 0.0  & 0.0  & 0.2   & 46.6 & 72.8         & \textbf{79.1} & 65.6$\pm$5.1              \\
antmaze-large-play-v2       & 0.0  & 6.7  & 0.0  & 0.0   & 16.4 & 42.0         & 35.3          & \textbf{42.4}$\pm$6.6     \\
antmaze-large-diverse-v2    & 0.0  & 2.2  & 0.0  & 0.0   & 3.2  & \textbf{56.0}& 36.3          & 39.2$\pm$8.9              \\
\hline
AntMaze sum                & 91.6 & 142.8& 95.4 & 120.2 & 218.0& \textbf{378.6}& 359.6          & 374.3$\pm$44.2          \\
\toprule[1pt]
\end{tabular}
}
\label{tb:result_mujoco}
\end{table*}

After training a diffusion model, we can apply the actions sampled from the learned policy to Eq. \ref{eq:diffusion_loss}, and approximate the $\log\pi_{\beta}(a|s)$.
Considering that the indicator function $\mathbb{I}(x)$ is difficult to train, we use the sigmoid function $\sigma(x)$ instead.
Combining the two parts of support constraint and density estimator, the loss function in Eq. \ref{eq:ploss1}
can be implemented as follows:
\begin{equation}\label{eq:ploss2}
    \begin{aligned}
        L_{\pi}(\theta)=\mathbb{E}_{s\sim D,a\sim \pi_{\theta}(\cdot|s)}[-Q_{\psi}(s,a)-\lambda\sigma[\alpha(\breve{\epsilon}-\mathcal{F}(a|s;\phi))]],
    \end{aligned}
\end{equation}
where $\breve{\epsilon}=-\hat{\epsilon}$, $\alpha$ is to scale the $\sigma(x)$ to be close to the indicator function $\mathbb{I}(x)$

We use TD3 \cite{td3} as our base algorithm, then the critic’s optimization objective is Eq \ref{eq:qloss}.
Our algorithm first trains the diffusion model using $L_{\pi_{\beta}}(\phi)$ to obtain a density estimator of $\pi_{\beta}(\cdot|s)$. 
Then we plug the regularization term computed by the diffusion density estimator into the policy optimation object $L_{\pi}(\theta)$ based on the Actor-Critic framework.

\begin{figure}[t]
	\centering
	\subfigure[halfcheetah]{
		\begin{minipage}[t]{0.22\linewidth}
			\centering
			\includegraphics[width=1\linewidth]{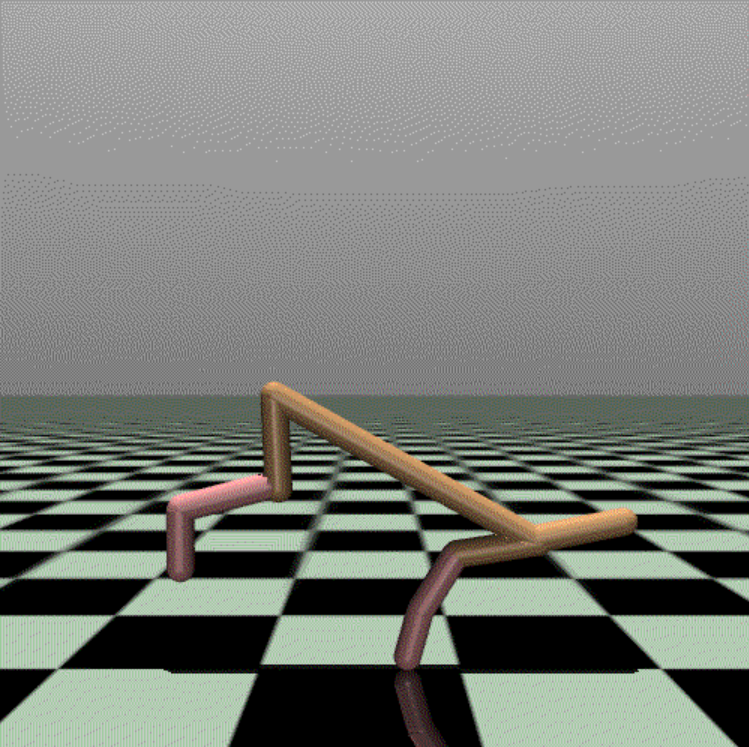}
		\end{minipage}
	}%
        \subfigure[hopper]{
		\begin{minipage}[t]{0.22\linewidth}
			\centering
			\includegraphics[width=1\linewidth]{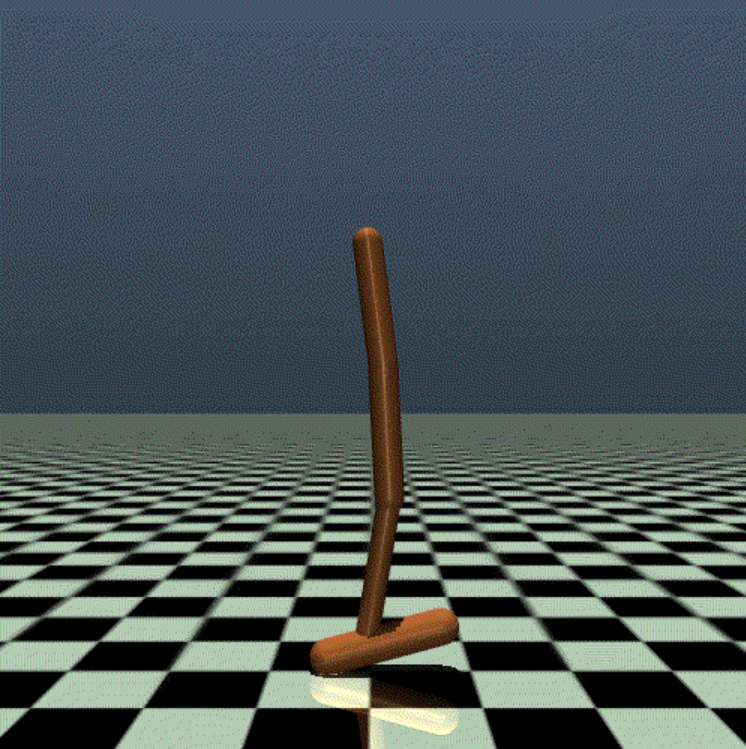}
		\end{minipage}
	}%
        \subfigure[walker2d]{
		\begin{minipage}[t]{0.22\linewidth}
			\centering
			\includegraphics[width=1\linewidth]{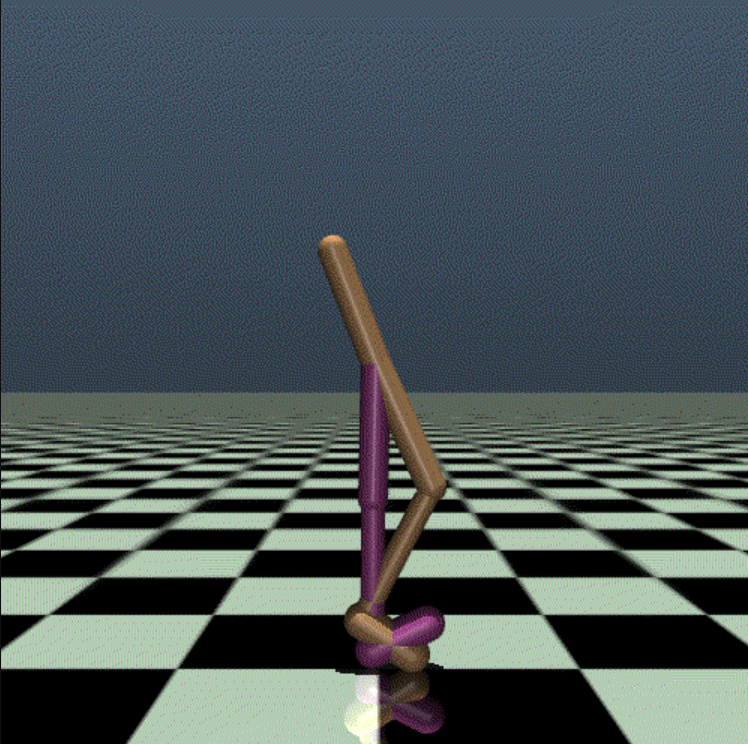}
		\end{minipage}
	}%
        \subfigure[antmaze]{
		\begin{minipage}[t]{0.22\linewidth}
			\centering
			\includegraphics[width=1\linewidth]{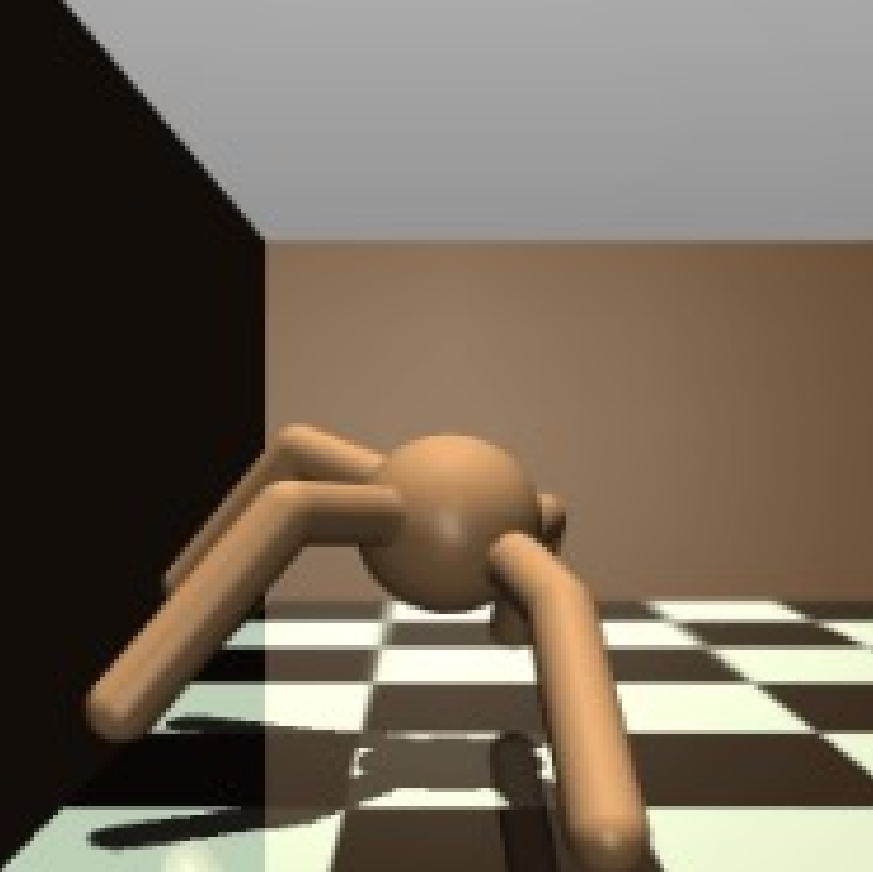}
		\end{minipage}
	}%
	\caption{Snapshots of tasks.}
	\label{fig:env}
\end{figure}

\section{Experiments}
\subsection{Task and Datasets}
We focus on evaluating our method on offline datasets provided by the D4RL benchmark ~\cite{d4rl}, including Gym-MuJoCo ~\cite{mujoco} and the AntMaze datasets. 
For Gym-MuJoCo, we choose halfcheetah, hopper, and walker2d as tasks. We use the three offline datasets including "medium", "medium-replay", and "medium-expert".
The AntMaze task is a challenging navigation scenario that needs control an 8-DoF ant quadruped robot to reach the goal location and receives a sparse 0-1 reward. We choose three different difficulty maps: umaze, medium, and large. In addition, each map includes a "play" task in which the goal is fixed and a "diverse" task in which the goal is variable.
Fig. \ref{fig:env} shows the snapshots of the halfcheetah, hopper, walker, and antmaze tasks.

\subsection{Baselines}
We compare OSC with prior state-of-the-art offline RL methods, including: BC \cite{bc}, BCQ \cite{bcq}, DT \cite{dt}, TD3+BC \cite{td3+bc}, CQL \cite{cql}, IQL \cite{iql}, and SPOT \cite{spot}. 
For the baseline, we directly report the normalized score from papers of prior methods or our replications.

\subsection{Performance Comparison on Offline RL}
The experimental results on the MuJoCo and AntMaze datasets are in Table ~\ref{tb:result_mujoco}. 
Notably, the OSC approach exhibited the highest average performance, surpassing all baseline methods in 6 out of 9 environments, and outperforming SPOT in 8 environments. 
Compared with other methods, in the suboptimal "medium" and "medium-replay" datasets, OSC obtains the highest performance. 
This arises from the fact that the action with the highest probability in the behavior policy is suboptimal, and the probability of the better action is low.
Our method does not impose extra constraints on actions within the support, OSC can freely choose better actions in the support, so the effect is the best.
In the "medium-expert" datasets, the optimal action and the action with the highest probability are relatively close.
While OSC did not achieve the highest performance, its performance still surpasses the previous constraint method SPOT. 

On the AntMaze datasets, OSC's average normalized score is slightly worse than IQL but better than other baselines including SPOT. This demonstrates that our method is a superior support constraint method.
Overall, this demonstrates the advantages of our method, which removes conservatism well.

\begin{table}[t]
\centering
\caption{
    The normalized scores of online fine-tuning after offline training on AntMaze datasets. 
    All experiments are the normalized scores of 1M steps of fine-tuning after offline training.
    For OSC, we report the mean and standard deviation for 8 seeds.
}
\resizebox{1\linewidth}{!}{
\begin{tabular}{l||c|c|c}
\toprule[1pt]
Dataset                     & IQL               & SPOT              & OSC(Ours)                        \\
\hline
antmaze-umaze-v2            & 85.4$\to$96.2     & 93.2$\to$99.2     & 95.3$\to$\textbf{99.5}$\pm$0.5    \\
antmaze-umaze-diverse-v2    & 70.8$\to$62.2     & 41.6$\to$96.0     & 60.2$\to$\textbf{98.3}$\pm$1.1    \\
antmaze-medium-play-v2      & 68.6$\to$89.8     & 75.2$\to$97.4     & 81.4$\to$\textbf{98.5}$\pm$2.2    \\
antmaze-medium-diverse-v2   & 73.4$\to$90.2     & 73.0$\to$96.2     & 63.3$\to$\textbf{97.9}$\pm$0.8    \\
antmaze-large-play-v2       & 40.0$\to$78.6     & 40.8$\to$89.4     & 42.5$\to$\textbf{90.8}$\pm$3.8    \\
antmaze-large-diverse-v2    & 40.4$\to$73.4     & 44.0$\to$90.8     & 21.2$\to$\textbf{91.9}$\pm$5.2    \\
\hline
AntMaze sum                 & 378.6$\to$490.4   & 367.8$\to$569.0   & 363.9$\to$\textbf{576.9}$\pm$13.6    \\
\toprule[1pt]
\end{tabular}
}
\label{tb:finetune}
\end{table}

\begin{figure*}[tb]
    \centering
    \includegraphics[width=0.85\linewidth]{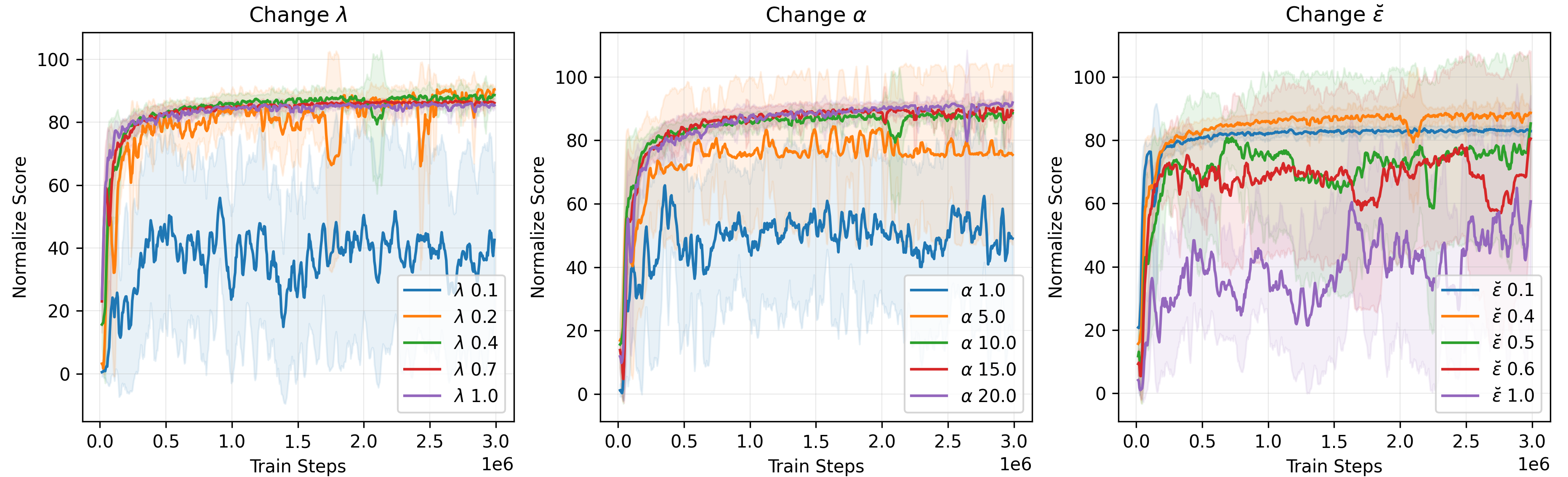}
    \caption{
    Analyze the impact of hyperparameters on the performance of OSC on the walker2d-medium datasets.
    \textbf{Left}: With varying values of hyperparameter $\lambda$, OSC applies support constraint with different strengths.
    \textbf{Middle}: As $\alpha$ changes, the degree to which the $\sigma(x)$ function is close to the indicator function $\mathbb{I}(x)$.
    \textbf{Right}: Different $\breve{\epsilon}$ represents different defined support bounds $\{a\in A|-\log\mu(a|s)<\breve{\epsilon}\}$
    }
    \label{fig:main_hyper}
\end{figure*}

\subsection{Online Fine-tuning after Offline RL}
The OSC method is very suitable for fine-tuning after offline RL training. 
Throughout the fine-tuning procedure, we gradually lessen the constraint strength $\lambda$ to progressively mitigate the conservatism present in the online training phase.
We compare our results with the IQL and SPOT algorithms on the AntMaze datasets, and the experimental results are shown in Table ~\ref{tb:finetune}.
Across all AntMaze datasets, the fine-tuning results of OSC outperform both of these methods.


\subsection{Ablation.}
\begin{figure}[t]
    \centering
    \includegraphics[width=1\linewidth]{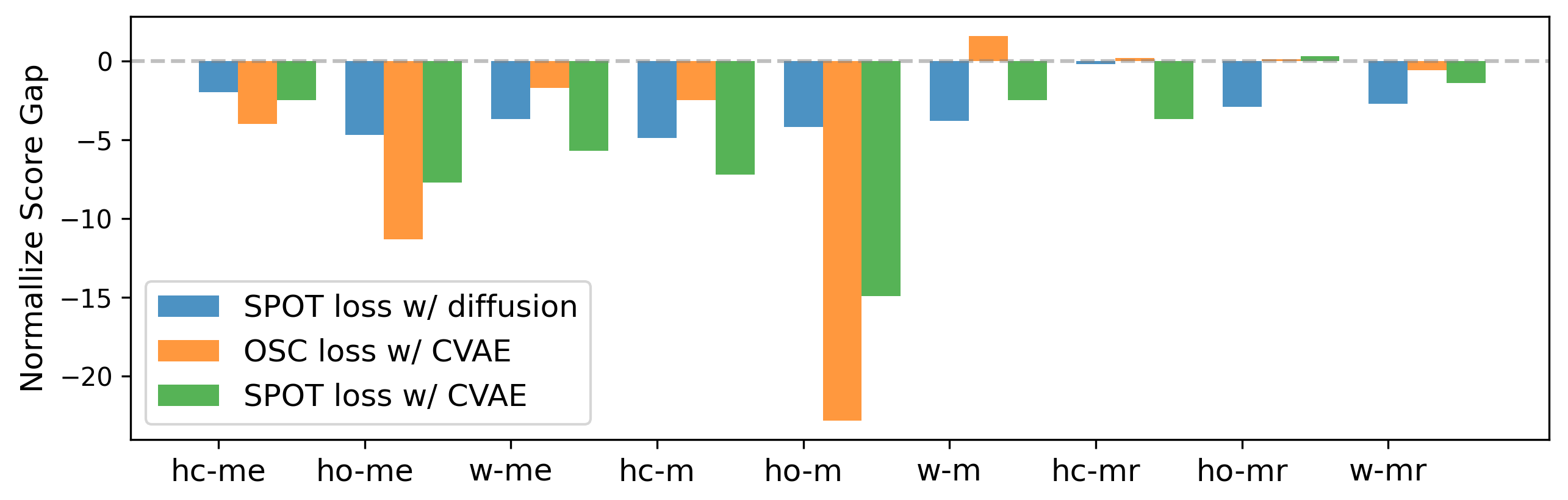}
    \caption{
    Degradation in normalized score of ablation methods, compared with the OSC. 
    OSC was compared with the following ablation methods: ablation of our proposed loss function, using SPOT loss function and diffusion density estimator; ablation of diffusion density estimator, using OSC loss and CVAE estimator; simultaneous ablation of OSC loss and diffusion estimator
    which is using SPOT loss and CVAE.
    hc=HalfCheetah, ho=Hopper, w=Walker2d, me=medium-expert, m=medium and mr=medium-replay.
    }
    \label{fig:ablation}
\end{figure}

\paragraph{Method ablation.}
As shown in Fig. \ref{fig:ablation}, we evaluate an ablation study over the components within our method. 
The majority of ablation methods perform worse than OSC.
Across most environments, the SPOT loss, utilizing the diffusion density estimator, exhibits superior performance compared to the SPOT which uses the CVAE density estimator. 
This shows the higher accuracy of diffusion-based density estimation for the behavior policy $\pi_{\beta}$.
Compared OSC with the SPOT loss that uses diffusion estimator, OSC outperforms in all environments, indicating that OSC better eliminates conservatism.
The combination of OSC loss and the CVAE density estimator, however, is not ideal across many environments.
As shown in Fig. \ref{fig:penalty}, this is attributed to the mutation of our loss constraint term near the support boundary $\hat{\epsilon}$, which necessitates a more precise support estimator.
Insufficiently accurate estimators struggle to fully realize the potential of OSC's loss.
Consequently, only when OSC loss and diffusion are used in conjunction, the full potential of OSC loss be realized, leading to optimal performance.
The ablation experiments of OSC validate the effectiveness of our method.

\paragraph{Hyperparameters influence.}
As shown in Fig. \ref{fig:main_hyper}, we illustrate the impact of three different hyperparameters which are $\lambda, \alpha, \breve{\epsilon}$.
For varying values of $\lambda$, the performance with small $\lambda$ is poor due to the small constraint term outside the support being unable to rigorously confine the learned policy within the support.
On the other hand, when $\lambda$ is relatively large, the performance differences between various $\lambda$ are minimal. This occurs because OSC is no constraint within the support, and changes in larger $\lambda$ do not affect that the learned policy selects optimal action within the support.
In the middle graph, as $\alpha$ increases, the performance improves, and the performance differences among different high alpha values are marginal.
This is attributed to the fact that a higher alpha allows the sigmoid function $\sigma(x)$ to approximate the ideal indicator function $\mathbb{I}(x)$ more closely. 
When the $\alpha$ is relatively large, the $\sigma(x)$ is close enough, so it is difficult to continue increasing performance.
Lastly, concerning the $\breve{\epsilon}$, a moderate support boundary $\breve{\epsilon}$ must be chosen for optimal effects.
If $\breve{\epsilon}$ is too large, the support includes many actions with low probabilities, leading to greater errors in the $Q$-function within the action space.
While if $\breve{\epsilon}$ is too small, it excludes potentially optimal actions from the support.
Overall, the hyperparameter experiments align with the characteristics of OSC, confirming the robustness of OSC  parameters $\lambda$ and $\alpha$, as well as highlighting the significance of the $\epsilon$.

\section{Conclusion}
We introduce the OSC, a novel support constraint method for offline RL. 
OSC introduces a support constraint term derived from the probability of learned policy within the support of behavior policy, enabling the learned policy to be confined within the support while not imposing constraints within the support.
This constraint term allows the policy to freely select optimal actions within the support.
Due to the nature of our constraint term, a more accurate estimation of the support is essential. Therefore, we utilize the diffusion model to characterize the density of the behavior policy.
We assess the performance of the OSC method on the D4RL benchmark, encompassing datasets such as MuJoCo and Antmaze, and our results surpass those of previous methodologies. 
This proves the effectiveness of our method.










\bibliographystyle{IEEEtran}
\balance
\bibliography{IEEEabrv,ref}

\end{document}